\documentclass[letterpaper]{article} % DO NOT CHANGE THIS
\usepackage[submission]{m3cs}  % DO NOT CHANGE THIS
\usepackage{times}  % DO NOT CHANGE THIS
\usepackage{helvet}  % DO NOT CHANGE THIS
\usepackage{courier}  % DO NOT CHANGE THIS
\usepackage[hyphens]{url}  % DO NOT CHANGE THIS
\usepackage{graphicx} % DO NOT CHANGE THIS
\usepackage{amssymb}
\usepackage{amsmath}
\usepackage{url}
\usepackage{color}
\usepackage{multirow}
\usepackage{subfigure}
\usepackage{array}
\usepackage{booktabs}
\usepackage{natbib}  % DO NOT CHANGE THIS AND DO NOT ADD ANY OPTIONS TO IT
\usepackage{caption} % DO NOT CHANGE THIS AND DO NOT ADD ANY OPTIONS TO IT
\usepackage{algorithm}
\usepackage{algorithmic}
\usepackage{bbding}
\usepackage{pifont}
\usepackage{newfloat}
\usepackage{listings}
\DeclareCaptionStyle{ruled}{labelfont=normalfont,labelsep=colon,strut=off} % DO NOT CHANGE THIS
\floatstyle{ruled}
\newfloat{listing}{tb}{lst}{}
\floatname{listing}{Listing}
\title{M$^3$CS: Multi-Target Masked Point Modeling with Learnable Codebook and Siamese
Decoders}
\author{
Qibo Qiu\textsuperscript{\rm 1,2}\equalcontrib,
Honghui Yang\textsuperscript{\rm 1}\equalcontrib,
Wenxiao Wang\textsuperscript{\rm 1},
Shun Zhang\textsuperscript{\rm 2},\\
Haiming Gao\textsuperscript{\rm 2},
Haochao Ying\textsuperscript{\rm 1},
Wei Hua\textsuperscript{\rm 2},
Xiaofei He\textsuperscript{\rm 1}
}

\affiliations{
	\textsuperscript{\rm 1} Zhejiang University \\
	\textsuperscript{\rm 2} Zhejiang Lab
}
\usepackage{bibentry}
\newcommand{\ie}{{\emph{i.e.}}}
\newcommand{\eg}{{\emph{e.g.}}}
\newcommand{\etc}{{\emph{etc.}}}

\begin{document}
\maketitle
\begin{abstract}
	\begin{quote}
		Masked point modeling has become a promising scheme of self-supervised pre-training for point clouds.
		Existing methods reconstruct either the original points or related features as the objective of pre-training.
		However, considering the diversity of downstream tasks, it is necessary for the model to have both low- and high-level representation modeling capabilities to capture geometric details and semantic contexts during pre-training.
		To this end, M$^3$CS is proposed to enable the model with the above abilities.
		Specifically, with masked point cloud as input, M$^3$CS introduces two decoders to predict masked representations and the original points simultaneously. While an extra decoder doubles parameters for the decoding process and may lead to overfitting, we propose siamese decoders to keep the amount of learnable parameters unchanged.
		Further, we propose an online codebook projecting continuous tokens into discrete ones before reconstructing masked points. In such way, we can enforce the decoder to take effect through the combinations of tokens rather than remembering each token.
		Comprehensive experiments show that M$^3$CS achieves superior performance at both classification and segmentation tasks, outperforming existing methods. 
	\end{quote}
\end{abstract}

\noindent Many applications, like mobile robotics, rely heavily on lidar point cloud-based 3D comprehension tasks, such as segmentation and classification. Recent advances in point cloud-based techniques can be attributed to the increase of labeled data. However, accumulating and annotating point clouds are laborious and time-consuming.
Motivated by masked image/language modeling methods (\eg, MAE, data2vec, BERT, \etc), researchers introduced Masked Point Modeling (MPM) pre-training~\cite{point-mae,point-bert,pointm2ae,voxel-mae,gdmae,mvjar,geomae} to alleviate the above problems.
It enables accurate representations of point clouds through MPM pre-training and demonstrates excellent transferability for downstream tasks.
Consequently, massive unlabeled data can be utilized and less task-specific labeled training data is required for consistent accuracy.
\begin{figure}[t]
		\flushleft
		\subfigure[\small Point-MAE, Point-M2AE, \etc]{\includegraphics[width=7.8cm]{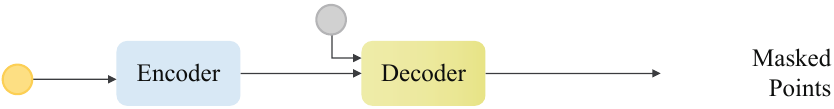}}
		\subfigure[\small Point2Vec, POS-BERT, \etc]{\includegraphics[width=7.8cm]{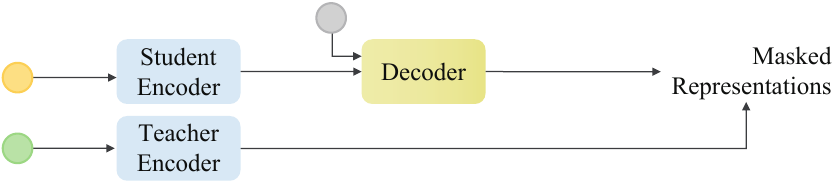}}
		\subfigure[\small Our vinalla framework with a simple combination.]{\includegraphics[width=7.8cm]{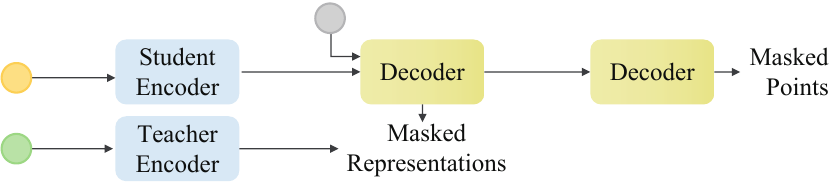}}
		\subfigure[\small M$^3$CS.]{\includegraphics[width=7.8cm]{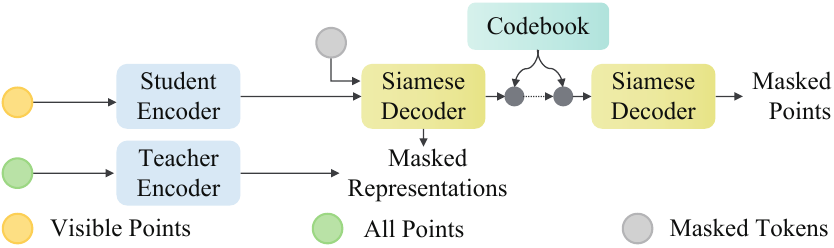}}
	\setlength{\abovecaptionskip}{0cm}
	\setlength{\belowcaptionskip}{0cm}
	\caption{Comparison of (a) point reconstruction, (b) representation reconstruction, (c) simple multi-target reconstruction, and (d) multi-target reconstruction with a learnable codebook and siamese decoders.}
	\label{algos}
	\vspace{-10pt}
\end{figure}

As shown in Figure~\ref{algos}(a), some existing MPM methods, such as Point-MAE~\cite{point-mae} and Point-M2AE~\cite{pointm2ae}, are proposed to reconstruct original masked points for pre-training. While Point2Vec~\cite{point2vec} in Figure~\ref{algos}(b) assumes that MPM prefers high-level representations to low-level\footnote{In this paper, we refer to high-level representations as those with sufficient semantics and low-level ones as those with original geometry structures (\eg, key points, normals, lines \etc)} ones (\eg, original points). Thus, similar to data2vec~\cite{data2vec}, Point2Vec resorts to representation reconstruction, enabling the model to capture high-level semantic information. 

Despite the  success of previous methods, we argue that both high-level representations and their low-level counterparts are essential to adapt the pre-trained model to various downstream tasks~\cite{pcl}, 
while previous methods only consider one of them. To this end, we propose a multi-target masked point modeling framework, and the vanilla design is as shown in Figure~\ref{algos}(c). That is, we employ representation and point reconstruction with two decoders, respectively, which is similar to the combination of Point-MAE and Point2Vec. In this way, the encoder is enforced to retain both low- and high-level representations. However, the additional decoder leads to higher model capacities and makes it easier to overfit, especially when datasets for point clouds are small with limited diversity.

To tackle this problem, we further introduce a codebook and siamese decoders into our multi-target masked point modeling framework. As shown in Figure~\ref{algos}(d), for the point reconstruction, the tokens with continuous representations are first projected into discrete tokens from the codebook. Through restricting the size of the codebook, we enforce the decoder to reconstruct masked points with semantic combinations of discrete tokens rather than \emph{remembering} the specific projection from a continuous token to its corresponding points. 
As an extra gift, the introduced codebook also benefits downstream tasks. 
\ie, when handling downstream tasks, we concatenate the continuous tokens with their discrete counterparts from the codebook to augment the representations for better performance.

Moreover, to further mitigate overfitting, we propose to carry out two decoding tasks with siamese decoders, \ie, sharing the architecture and weights for both decoders. This special design counteracts the superfluous capacities introduced by an extra decoder. This is motivated by that plenty of previous papers have proven that one single Transformer-based model may qualify various tasks with different prompts.
And our siamese decoders fall into this paradigm, \ie, the decoder takes the masked tokens as ``prompts'' to finish representation reconstruction, while taking representations as ``prompts'' to reconstruct the masked points.
The contributions can be summarized as follows:
\begin{itemize}
	\item We propose a multi-target masked modeling framework for point cloud Transformer pre-training, enabling the encoder with both low- and high-level representations.
	\item A learnable codebook is introduced to discretize tokens, which not only mitigates the overfitting during the pre-training stage, but also facilitates the downstream tasks with extra semantic representations.
	\item We propose siamese decoders to counteract the superfluous capacities introduced by the extra decoder, which further mitigates the decoder from overfitting and also improves the pre-training performance.
	\item Plenty of experiments are conducted on various downstream tasks and show the effectiveness of our proposed method, as well as the delicate components.
\end{itemize}
\section{Related Work}
\textbf{Masked Point Modeling. } Informed by the success of Masked Image Modeling (MIM), typical methods attempt to perform the pretext task to reconstruct the masked points from the visible parts. 3D representation can be learned from the massive unlabeled data in this manner. Point-BERT \cite{point-bert} is the first to introduce BERT \cite{bert} for MPM, in which point clouds can be processed similarly with language and images by embedding point patches into discrete tokens. Nonetheless, the tokenizer must be trained offline by a discrete Variational AutoEncoder (dVAE) \cite{dvae}, making the method a complex two-stage solution, and the tokenizer cannot be modified during subsequent fine-tuning. To this end, Point-MAE \cite{point-mae} proposes a novel scheme entirely based on the standard Transformer to address the complexity of architecture and early leakage of location information. However, it has no explicit high-level representation learning strategy, but leverages the capability of long-range statistical dependencies learning in origin MAE \cite{understand_mae}. POS-BERT~\cite{pos-bert} and Point2Vec~\cite{point2vec} perform explicitly high-level representation learning  by introducing a contrastive learning strategy, and the contrastive views are constructed by different data augmentations, while these methods have no masked point reconstruction process.

Considering that point clouds consist of simple and detailed geometric elements with strong generalization, point reconstruction is still needed for low-level representation learning during pre-training. M$^3$CS provides a novel learning pipeline for both low- and high-level representations.

\textbf{Codebook for Representation. }
Since there is no mature BPE \cite{bpe} algorithm to divide point clouds or images into meaningful patches. A straightforward method is to divide an image into fixed-size patches, and project them into continuous tokens by a follow-up MLP layer in the ViT-style \cite{vit}. VQ-VAE \cite{vqvae} provides a novel method that further projects the continuous tokens into discrete token space. However, the operation of the nearest neighbor lookup makes it hard to train. To this end, DALLE~\cite{dalle} introduces the Gumbel-softmax \cite{gumbel} relaxation for vector quantization, which is a replacement for the nearest neighbor lookup and is friendly to train. Following researches in computer vision have demonstrated that learning meaningful discrete tokens (codebook) through vector quantization is an effective technique for image generation. Inspired by this, BEIT \cite{beit} adapts a codebook for BERT-style MIM and obtains superior performance. While the codebook in BEIT is provided by a trained tokenizer, which makes BEIT a complex two-stage method. In addition, the improved version BEITv2 \cite{beitv2} introduces the Vector-Quantization Knowledge Distillation (VQ-KD) algorithm to learn a codebook online, and additional analysis shows that the codebook is meaningful.

Point-BERT is the pioneering MPM work that makes use of a codebook trained by dVAE, which is inspired by BEIT. In this paper, we use a codebook for robust and accurate masked point reconstruction, and it can be learned online. In addition, M$^3$CS makes innovative use of the codebook not only in pre-training but also in downstream fine-tuning.

\begin{figure*}[!h]
	\begin{center}
		\includegraphics[width=17.5cm]{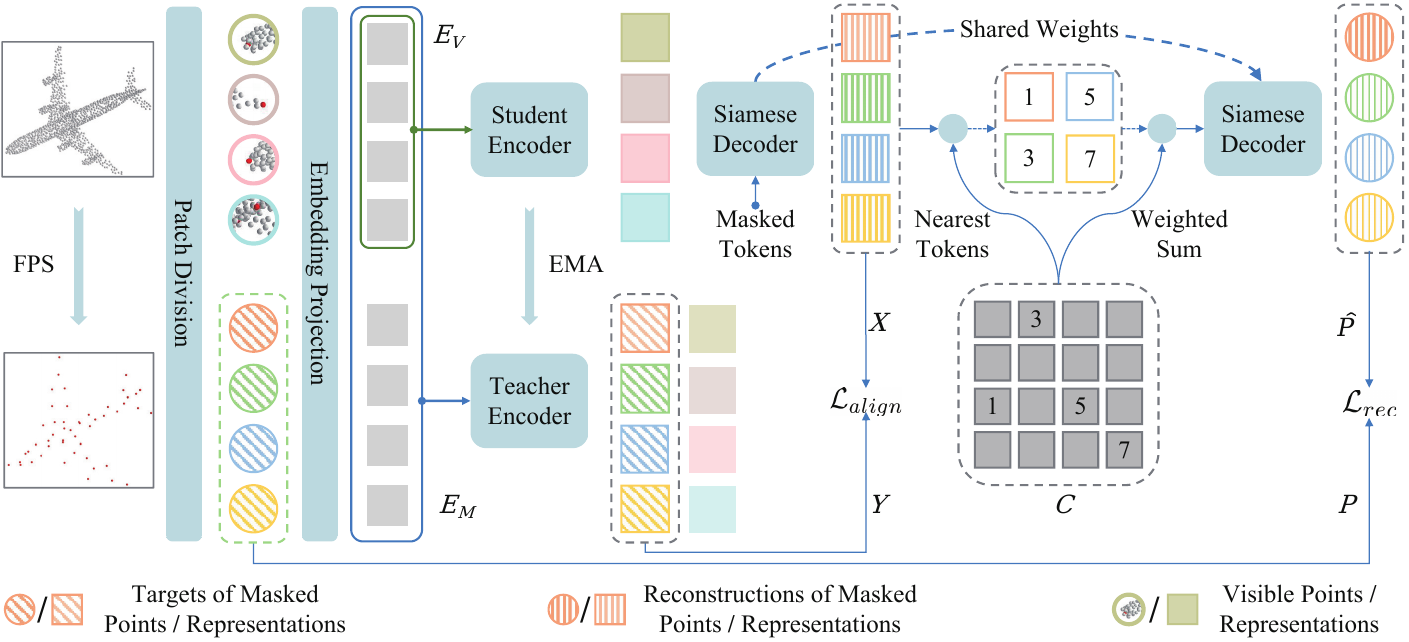}
	\end{center}
	\caption{Pre-training pipeline of M$^3$CS. The point cloud is divided into patches based on the FPS-sampled centers, then both visible and masked patches are projected into continuous tokens. The student and teacher encoder take $E_V$ and $Cat(E_V, E_M)$ as input, respectively. The representation decoder is intended to predict the mask representations, while the point decoder reconstructs the masked points. Moreover, these decoders are siamese ones, sharing both weights and the architecture. In addition, masked points are reconstructed based on combinations of meaningful discrete tokens in the codebook $C$.}
	\label{arch}
	\vspace{-10pt}
\end{figure*}

\section{Method}

Our method contains pre-training and fine-tuning procedures, which will be discussed in detail in the following two subsections, respectively. The pre-training process employs a multi-target pretext task with a learnable codebook and siamese decoders. As for fine-tuning, we will highlight our feature aggregation methods for downstream tasks.
\subsection{Multi-Target Pre-Training}
The overall pre-training process is shown in Figure~\ref{arch}. To ensure that point clouds can be processed by standard Transformer similarly to language and images. We divide a point cloud $P$ into $G$ patches following Point-MAE, and the center of each patch is generated by the Farthest Point Sampling (FPS) algorithm once $G$ is determined. A center and its $k$ nearest neighbors obtained by the $k$-NN algorithm constitute a patch. Given a patch  $P_i \in \mathbb{R}^{(1+k)\times3}$,  we use a neat and lightweight mini-PointNet with  max pooling \cite{point2vec}  to project it into a continuous token $E_i \in \mathbb{R}^{1\times c}$ following the principle of permutation invariance.

Then, the tokens are simultaneously fed into the student and teacher encoders.
The student encoder only takes the visible patches as inputs to learn holistic representations of point clouds, and thereafter reconstructs the masked representations from the visible counterparts.
Instead, the teacher encoder accepts all uncorrupted patches as inputs to extract contextualized representations from a complete view, thereby providing high-level semantic supervision for the student encoder.
Considering that point clouds consist of detailed geometric elements with strong generalization, it is also necessary to learn the representation of low-level visual clues.
Instead of directly using the reconstructed representations for masked point recovery, we resort to predicting points from a combination of discrete tokens in an online learning codebook.
Finally, weight-shared siamese decoders are used for both the low- and high-level reconstructions to achieve comprehensive representation learning.
In the following sections, we will detail the two targets of representation reconstruction and point reconstruction to facilitate the capture of high- and low-level representations, respectively.

\textbf{Target 1: Representation Reconstruction. }
With the partial inputs of the student encoder and the complete inputs of the teacher encoder, the target of masked semantic features $Y$ can be explicitly aligned with the corresponding reconstructed representation $X$ as follows: 
\begin{align}
	& X = Decoder(StudentEncoder(E_V)),\\
	& Y =TeacherEncoder(Cat(E_V, E_M)), \\
	& \mathcal{L}_{align} = Smooth\text{-}{\ell_1}(X, Y),
\end{align}
where $E_V$ and $E_M$ denote the continuous visible and masked tokens, respectively. The representation decoder is intended to predict masked representations based on visible counterparts, and only representations at the masked positions will be aligned by the \textit{Smooth}-$\ell_1$ loss.

To obtain a stable and accurate teacher encoder, we adapt the Exponential Moving Average (EMA) updating strategy~\cite{meanteacher}. Specifically, the weight of the teacher encoder will be updated as follows:
\begin{equation}
	\theta_s=\lambda\theta_{s-1}+(1-\lambda)\theta_e,
\end{equation}
where $\theta_s$ and $\theta_{s-1}$ are the weights of teacher encoder in $s$-th and ($s$-1)-th train step, respectively. $\theta_e$ is the weight of student encoder in ($s$-1)-th train step, and $\lambda\in[0,1)$ is a momentum parameter scheduled dynamically during training.

\textbf{Target 2: Point Reconstruction with Learnable Codebook. }
The masked and visible patches share a portion of high-level semantics in a point cloud frame, which can be learned by a MAE-style point reconstruction \cite{point-mae}. In contrast to reconstructing points directly based on the prediction of masked representations, we predict the coordinates of masked points after projecting corresponding continuous tokens into meaningful discrete space as follows:
\begin{align}
	& Z = GumbelSoftmax(Linear(X)),\\
	& F = Decoder(Z \times C^T), \\
	& \widehat{P} = Linear(F),
\end{align}
where $X$ indicates the masked representations. A linear layer is intended to predict the distribution over discrete tokens (codebook), which is a differentiable replacement of the nearest token lookup. Moreover, Gumbel-softmax relaxation is further employed to normalize the distribution $Z$. The point decoder generates rebuilding features based on the weighted sum of discrete tokens in codebook, and the coordinates of masked points are predicted by a linear layer. Specifically, given a point cloud $P$, if the $i$-th patch $P_i$ is masked, we calculate the Chamfer Distance~\cite{cdloss} between $P_i$ and its prediction $\widehat{P}_i$ as follows:
\begin{equation}
	\mathcal{L}_{rec}=\frac{1}{\left|\widehat{P}_i\right|}\sum_{p\in \widehat{P}_i}\min_{q\in P_{i}}\left\|p-q\right\|_2^2+\frac{1}{\left|P_{i}\right|}\sum_{q\in P_{i}}\min_{p\in \widehat{P}_i}\left\|p-q\right\|_2^2.
\end{equation}
The codebook  $C = \left\{C_0, C_1, \ldots, C_T\right\} \in \mathbb{R}^{T \times D}$ can be learned during point reconstruction, each discrete token $C_t$ with $t \in[1,2,\ldots,T]$ can be seen as a semantic centroid. Meaningful combinations of semantic centroids can contribute to accurate masked points recovery, and those centroids can be further adapted in downstream fine-tuning.

\textbf{Siamese Decoders. }
We propose siamese decoders to mitigate overfitting in pre-training. Specifically, the representation and point decoders consist of the same blocks and shared weights, thus, less parameters are required for pre-training. Each decoder consists of $4$ self-attention blocks, and positional embeddings are added to every block. 
Only the inputs of siamese decoders are different, which can be treated as various ``prompts'' to decode multiple targets. 
Since two targets are decoded in the same self-attention space, the decoders will maximize the commonality of two decoding processes. The siamese design can prevent the decoders from overfitting the two objectives independently when pursuing the minimum total loss. Therefore, it makes the low- and high-level representations generated by encoder more versatile, while achieving the same pre-training objective. The siamese design helps to exploit the capability of encoder, and the performance of fine-tuning will be enhanced.

\textbf{Total Loss. }
M$^3$CS employs two targets to supervise the pretext task for pre-training as follows:
\begin{equation}
	\mathcal{L}_{total}=\mathcal{L}_{align}+\eta \mathcal{L}_{rec},
\end{equation}
where $\mathcal{L}_{total}$ is the training loss for the pretext task, $\mathcal{L}_{align}$ and $\mathcal{L}_{rec}$ denote the loss for representation and point reconstruction, respectively. $\eta$ is intended to balance two targets, which is set to $1$ in our experiments. 

\subsection{Fine-Tuning for Downstream Tasks}
\begin{figure}[t]
	\begin{center}
		\includegraphics[width=8cm]{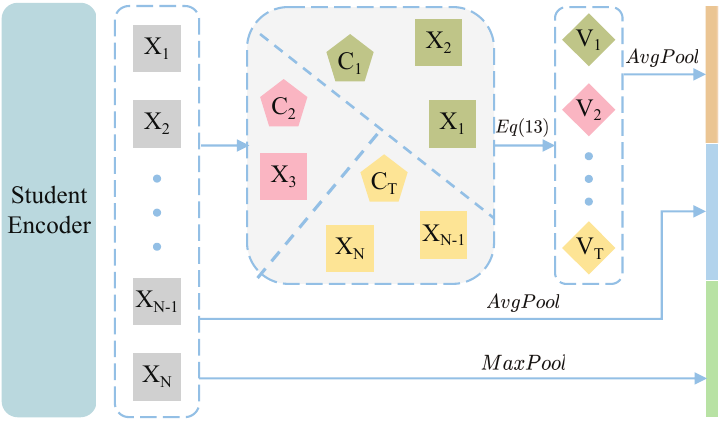}
	\end{center}
	\caption{Hybrid Token Aggregation (HTA). Typical methods \cite{point-mae,point2vec} generate a global representation by performing max- or average-pooling operations on continuous tokens, which is utilized in downstream tasks. Differently, this paper employs HTA strategy for semantically meaningful global representation.}
	\label{vlad}
	\vspace{-10pt}
\end{figure}

After pretext task pre-training, the model is fine-tuned for downstream tasks. During the fine-tuning process, previous works often compute max- and average-pooling of continuous tokens to generate global representations for downstream tasks. Contrastly, motivated by our pre-training setting that a learnable codebook naturally forms many semantic centroids, we believe incorporating these centroids into global representations will also benefit downstream tasks. To this end, as shown in Figure~\ref{vlad}, we propose a Hybrid Token Aggregation (HTA) strategy. \ie, HTA formulates the continuous tokens before the task-specific head as:
\begin{equation}\small
	O = Concat(AvgPool(X), MaxPool(X), {\cal G}(X, C)),
\end{equation}
where $O$ is the representation for a downstream task, and ${\cal G}(X, C)$ represents the global representation for all centroids in codebook $C$. It is generated by operating average-pooling on the statistic embeddings set $V$ as follows:
\begin{align}
	& {\cal G}(X, C) = AvgPool(V, dim=0), \\
	& V = \left \{V_0, V_1, \ldots, V_T\right \} \in \mathbb{R}^{T \times D},
\end{align}
where $V_t$ indicates the statistic embedding for the $t$-th centroid in the codebook. Here, we employ NetVLAD~\cite{netvlad} to calculate the statistic embedding.
In particular, it generates a statistic embedding for each centroid in the codebook by accumulating the distances between each continuous token and the centroid as follows:
\begin{align}
	& V_t = \sum_{t=1}^{N}{\alpha_t(X_j)(X_j - C_t)},\\
	& \alpha_t(X_j) = \frac{e^{W_t^T X_j+B_t}}{\sum_{t^{\prime}=1}^{T} e^{W_{t^{\prime}}^T X_j+B_{t^{\prime}}}},
\end{align}
where $C_t$ and $X_j$ denote the embedding of the $t$-th centroid in the codebook and the $j$-th continuous tokens, respectively. $T$ is the size of the codebook, and $\alpha_t(X_j)$ indicates the possibility that $X_j$ belongs to $C_t$. $W_t$ and $B_t$ are learnable parameters implemented by an MLP layer. Thus, the HTA pipeline can be trained in an end-to-end manner. 

It is worth noting that $X$ contains continuous tokens from specific hidden layers in the encoder.
Since different downstream tasks rely differentially on each hidden layer. same as the methods being compared, we fetch continuous tokens from hidden layers $\left \{11\right \}$ and $\left \{3,7,11\right \}$ for classification and segmentation tasks, respectively.

\section{Experiments}
In this section, we conduct both comparative experiments and ablation studies to verify the effectiveness of the proposed M$^3$CS. For a fair comparison, the pretext task is trained on the training set of ShapeNet \cite{shapenet}, which consists of nearly $51300$ synthetic models from $55$ object categories. The fine-tuning performance is tested on $4$ downstream tasks, including object classification on ScanObjectNN \cite{scanobject} and ModelNet40 \cite{modelnet40} dataset, few-shot classification on ModelNet40 dataset and part segmentation on ShapeNetPart \cite{part_shapenet} dataset. During the ablation studies,  effectiveness of point reconstruction based on codebook,  influence of siamese decoders and masking strategy are further explored. 
\subsection{Implementation Details}
In order to maintain compatibility with the previous outstanding works Point-BERT, Point-MAE and Point2Vec, this paper employs the same encoder, which consists of a standard Transformer with 12 self-attention blocks. Each of the siamese decoders has the same architecture as the decoder in Point2Vec. Position embedding in encoder and decoders is implemented by a shared learnable MLP, mapping the center of each patch into a $c$-dimension vector.
During pre-training, the number of input points is sampled at random to 1024 per frame, and these points are then grouped into 64 subsets of 32 points each. Since the implementation of M$^3$CS is based on the codebase of Point2Vec with minimal modifications, the hyper-parameters, including epochs, batch size, mask ratio, and data augmentations, remain identical to those of Point2Vec. We reproduce the results of Point2Vec in our environment as the $baseline$, as previously  mentioned\footnote{\textcolor[rgb]{0,0,0}{\url{https://github.com/kabouzeid/point2vec}}}, the reproducibility of Point2Vec is sensitive to both software and hardware environments, consequently, the results of Point2Vec differ from those in the original paper.
\begin{table}
	\setlength{\abovecaptionskip}{0cm}
	\renewcommand\arraystretch{0.8}
	\centering
	\caption{
		Object classification on the synthetic ModelNet40 dataset. Overall accuracy with and without voting strategy are reported, * denotes results in our environments.
	}
	\label{cls_modelnet}
	\begin{tabular}{lcc}
		\hline \hline
		\specialrule{0em}{1pt}{1pt}
		Method        & $+$Voting                 & $-$Voting   \\ \hline
		\specialrule{0em}{1pt}{1pt}
		Transformer-OcCo & 92.1 & --  \\
		OcCo & 93.0 & --  \\
		STRL & 93.1 & --  \\
		Point-BERT  & 93.2 & 92.7  \\
		ParAE & -- &  92.9 \\
		PointGLR & -- & 93.0 \\
		MaskPoint & 93.8 & --  \\
		Point-MAE & 93.8& 93.2  \\
		Point2Vec*   & 93.9 & 93.8 \\
		Point-M2AE & 94.0  & 93.4 \\
		PointGPT-S & 94.0 & - \\
		M$^3$CS* (Ours) & \textbf{94.3} &  \textbf{94.1}\\
		\hline \hline
	\end{tabular}
	\vspace{-10pt}
\end{table}
\begin{figure}[!ht]
	\begin{center}
		\includegraphics[width=8cm]{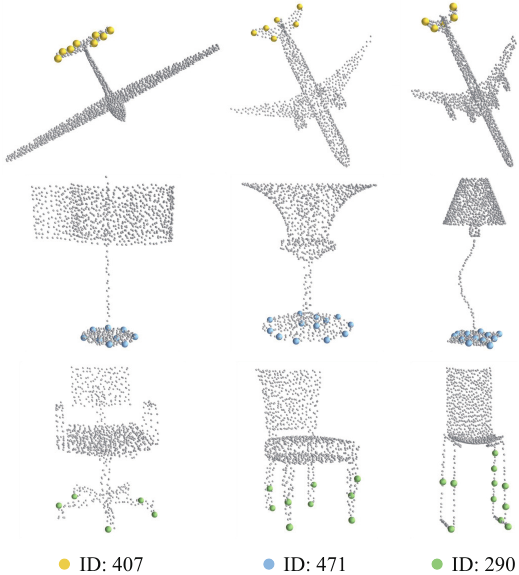}
	\end{center}
	\caption{Visualization of discrete tokens in codebook. Only center points belong to specific tokens are colored, and each token ID is represented by a unique color.}
	\label{tokens}
	\vspace{-10pt}
\end{figure}
\subsection{Fine-Tuning Results}
To evaluate the transferability of the proposed M$^3$CS, many popular methods including OcCo \cite{occo}, Transformer-OcCo \cite{point-bert}, ParAE \cite{pare}, STRL \cite{strl}, PointGLR \cite{pointglr}, Point-BERT, MaskPoint \cite{maskpoint}, PointMAE, Point-M2AE \cite{pointm2ae}, Point2Vec and PointGPT-S \cite{pointgpt} are adapted in the comparative experiments.

\textbf{Object Classification on a Synthetic Dataset. } To evaluate the classification performance, we conduct fine-tuning on the ModelNet40 dataset. Which contains $12311$ clean CAD models of $40$ categories. For a fair comparison, the dataset is divided into $9843/2468$ for training and testing, moreover, the model is pre-trained on ShapeNet and the voting strategy \cite{voting} is employed. According to Table \ref{cls_modelnet}, M$^3$CS obtains 94.3\% accuracy with voting, outperforming the previous best method by 0.3\% accuracy. As shown in Figure~\ref{tokens}, we further visualize the semantic centroids in the codebook fine-tuned on the ModelNet40 dataset, revealing the relationship between the discrete tokens and meaningful point cloud patches.

\textbf{Object Classification on a Real-world Dataset. }
\begin{table}[!htb]
	\setlength{\abovecaptionskip}{0cm}
	\renewcommand\arraystretch{0.8}
	\centering
	\caption{
		Classification on the real-world ScanObjectNN dataset,
		we report the overall accuracy of three settings. * denotes results in our environments.
	}
	\label{scanobjectnn}
	\begin{tabular}{lccc}
		\hline \hline
		\specialrule{0em}{1pt}{1pt}
		Method & \footnotesize  OBJ-BG & \footnotesize  OBJ-ONLY & \footnotesize  PB-T50-RS \\
		\hline
		\specialrule{0em}{1pt}{1pt}
		Transformer-OcCo & 84.9 & 85.5 & 78.8 \\
		Point-BERT  & 87.4 & 88.1 &  83.1                       \\
		MaskPoint & 89.3 & 89.7 & 84.6 \\
		Point-MAE & 90.0 & 88.3 &  85.2                  \\
		Point-M2AE & 91.2 & 88.8 &  86.4  \\
		PointGPT-S & 91.6 & 90.0 & 86.9 \\
		Point2Vec* & 91.7 & 90.0 & 87.1\\
		M$^3$CS* (Ours) & \textbf{91.9} & \textbf{90.7} & \textbf{87.6} \\
		\hline \hline
	\end{tabular}
	\vspace{-10pt}
\end{table}
Due to the variations in the real world, it is difficult for a self-supervised pre-trained model to transfer from synthetics to real-world data. In this paper, object classification performance is tested on the real-world ScanObjectNN dataset, which contains $2902$ scans of real-world objects from $15$ classes. 
Three variants with various levels of difficulty are tested. The OBJ-ONLY variant contains only ground truth objects, which more closely resembles synthetic data. While objects in the OBJ-BG variant are attached with background data. The PB-T50-RS is the hardest variant, where objects are randomly shifted up to $50\%$ of their sizes along the world axis and additional rotation/scaling is applied. Therefore, The PB-T50-RS variant consists of five random samples for each original object, totaling $14510$ perturbed scans.
As shown in Table \ref{scanobjectnn}, M$^3$CS outperforms the previous best method by 0.2\%, 0.7\% and 0.5\% on OBJ-BG, OBJ-ONLY and PB-T50-RS, respectively. It is notable that M$^3$CS only employs a standard Transformer as the backbone. While Point-M2AE leverages a multi-scale architecture to capture local representations and PointGPT-S proposes a GPT scheme with a dual masking strategy for self-supervised learning. Results in Table \ref{scanobjectnn} demonstrate that the neat standard Transformer trained by the M$^3$CS paradigm can generate well from synthetics to real-world data.
\begin{table*}[!htb]
	\setlength{\abovecaptionskip}{0cm}
	\renewcommand\arraystretch{0.8}
	\caption{Part segmentation results on ShapeNetPart. * denotes results in our environments. In contrast to the standard Transformer backbone, \textbf{MT} means more complicated Transformer-based backbones with corresponding complex masking strategies.}
\label{part_seg}
\centering
\begin{tabular}{lccccccccccccccccccc}
	\hline \hline
	\specialrule{0em}{1pt}{1pt}
	\multirow{2}{*}{Method}   & \multirow{2}{*}{MT}	& \multirow{2}{*}{$mIoU_C$}    &   aero &  bag  &  cap  &  car  &  chair &  earph &  guitar &  knife \\ & &&  lamp &  laptop &  motor &  mug  &  pistol &  rocket &  skateb &  table \\
	\hline
	\specialrule{0em}{1pt}{1pt}
	Transformer-OcCo &\ding{55}& 83.4   &  83.3 &  85.2 &  88.3 &  79.9 &  90.7  &  74.1  &  91.9   &  87.6  \\ & &&  84.7 &  95.4   &  75.5   &  94.4  &  84.1  &  63.1    &  75.7 &  80.8 \\
	Point-BERT &\ding{55}       & 84.1   &  84.3 &  84.8 &  88.0 &  79.8 &  91.0  &  81.7  &  91.6   &  87.9   \\ & &&  85.2 &  95.6   &  75.6  &  94.7   &  84.3   &  63.4   &  76.3 &  81.5  \\
	Point-MAE &\ding{55}   	 & 84.1   &  84.3 &  85.0 &  88.3 &  80.5 &  91.3  &  78.5  &  92.1   &  87.4   \\ & &&  86.1 &  96.1   &  75.2  &  94.6    &  84.7   &  63.5   &  77.1   &  82.4  \\
	PointGPT-S&\Checkmark  & 84.1   &  --   &  --   &  --   &  --   &  --    &  --    &  --     &  --     \\ & &&  --   &  --     &  --    &  --   &  --     &  --     &  --     &  --    \\
	Point2Vec* &\ding{55} & 84.2  &  84.6 &  85.7 &  87.6  &  80.5 &  91.5 &  77.9   &  91.9  &  88.0     \\ & &&  85.4 & 96.0   &  76.1  &  95.2    &  85.5     &  63.2  &  76.3 &  81.6    \\
	MaskPoint 	&\ding{55}	 & 84.4  &  84.2 &  85.6 &  88.1  &  80.3 &  91.2  &  79.5 &  91.9   &  87.8    \\ & &&  86.2 &  95.3  &  76.9  &  95.0   &  85.3   &  64.4   &  76.9 &  81.8 \\
	Point-M2AE &\Checkmark & \textbf{84.9} &  --   &  --   &  --  &  --   &  --    &  --    &  --     &  --     \\ & &&  --   &  --     &  --    &  --   &  --     &  --     &  --     &  --    \\
	M$^3$CS* (Ours)   &\ding{55}     & 84.6  &  85.0 & 85.5 &  88.5   &  81.0  &  91.4 &  80.9  &  91.7   &  87.7     \\ &&&  85.1 &  96.1    &  77.3  &  95.4   &  85.0     &  64.2  &  77.0  &  81.4    \\
	\hline \hline
	\vspace{-10pt}
\end{tabular}
\end{table*} 

\textbf{Part Segmentation. }
Beyond classification testing, we also evaluate the performance of semantic segmentation on the ShapeNetPart dataset, which includes $16880$ models from $16$ shape categories and $50$ different part categories. Moreover, the dataset is split $14006/2874$ for training and testing following a fair experiment setting.

In Table \ref{part_seg}, M$^3$CS outperforms Point2Vec and Point-MAE by 0.4\%  and 0.5\% $mIoU_C$, respectively. In addition, it surpasses the auto-regressive PointGPT-S by 0.5\% $mIoU_C$. Point-M2AE outperforms M$^3$CS mainly because the multi-scale Transformer-based backbone with a complex masking strategy has a large gain for segmentation tasks. 
Although experiments in Table \ref{part_seg} are not conducted to compare different backbone networks, we still present the results of Point-M2AE for a comprehensive discussion.

\textbf{Few-Shot Classification. }
Since the self-supervised pre-training pipeline is intended to capture the general representations for point cloud patches. In this paper, few-shot performance on ModelNet40 classification is tested. Following the “$K$-way $N$-shot” experimental setting, where $K$ classes are first selected at random, which is followed by the selection of $N + 20$ objects from each class. The model is trained by $K\times N$ samples (support set), and evaluated on the remaining $K \times 20$ samples (query set).

Table \ref{fewshot} shows that M$^3$CS outperforms the Point2Vec by significant margins, as evidenced by mean accuracy improvements of 0.7\% and 0.8\% in "5-way 10-shot" and "10-way 10-shot" tests, respectively. With the help of a robust and meaningful codebook, M$^3$CS outperforms other sophisticated methods especially in the more difficult 10-shot tests, achieving a new state-of-the-art performance. Moreover, M$^3$CS can generate more stable test results, as supported by smaller standard deviations.

\begin{table*}[!htb]
\setlength{\abovecaptionskip}{0cm}
\setlength{\belowcaptionskip}{0cm}
\renewcommand\arraystretch{0.8}
\caption{
	Few-shot classification results on ModelNet40, both mean accuracy and  its standard deviation are reported, calculating from $10$ independent runs. * denotes results in our environments.
}
\label{fewshot}
\centering
	\begin{tabular}{lcccccc}
		\hline \hline
		\specialrule{0em}{1pt}{1pt}
		Method           && 5-way 10-shot                                              &                             5-way 20-shot                &&                         10-way 10-shot                        &                     10-way  20-shot \\
		\hline
		\specialrule{0em}{1pt}{1pt}
		OcCo 				 && $91.9 $\footnotesize $\pm 3.6$ & $93.9 $\footnotesize $\pm 3.1$ && $86.4 $\footnotesize $\pm 5.4$ & $91.3 $\footnotesize $\pm 4.6$ \\
		Transformer-OcCo  && $94.0 $\footnotesize $\pm 3.6$ & $95.9 $\footnotesize $\pm 2.3$ && $89.4 $\footnotesize $\pm 5.1$ & $92.4 $\footnotesize $\pm 4.6$ \\
		Point-BERT	    && $94.6 $\footnotesize $\pm 3.1$  & $96.3 $\footnotesize $\pm 2.7$ && $91.0 $\footnotesize $\pm 5.4$  & $92.7 $\footnotesize $\pm 5.1$ \\
		MaskPoint		&& $95.0 $\footnotesize $\pm 3.7$ & $97.2 $\footnotesize $\pm 1.7$ && $91.4 $\footnotesize $\pm 4.0$ & $93.4 $\footnotesize $\pm 3.5$ \\
		Point-MAE      && $96.3 $\footnotesize $\pm 2.5$  & $97.8 $\footnotesize $\pm 1.8$ && $92.6 $\footnotesize $\pm 4.1$ & $95.0 $\footnotesize $\pm 3.0$ \\
		Point-M2AE	  && $96.8 $\footnotesize $\pm 1.8$   & $98.3 $\footnotesize $\pm 1.4$ && $92.3 $\footnotesize $\pm 4.5$ & $95.0 $\footnotesize $\pm 3.0$ \\
		PointGPT-S	&& $96.8 $\footnotesize $\pm 2.0$   & $98.6 $\footnotesize $\pm 1.1$ && $92.6 $\footnotesize $\pm 4.6$ & $95.2 $\footnotesize $\pm 3.4$ \\
		Point2Vec*                    && $96.7 $\footnotesize $\pm 2.6$           & $\mathbf{99.0} $\footnotesize $\pm 1.1$             && $93.7 $\footnotesize $\pm 4.1$             & $95.6 $\footnotesize $\pm 3.2$             \\
		
		M$^3$CS* (Ours)			&& $\mathbf{97.4} $\footnotesize $\pm 2.2$  & $98.9 $\footnotesize $\pm 1.2$    && $\mathbf{94.5} $\footnotesize $\pm 3.8$  & $\mathbf{95.7} $\footnotesize $\pm 2.9$    \\
		\hline \hline
	\end{tabular}
	\vspace{-10pt}
\end{table*}

\begin{table}[!htb]
	\renewcommand\arraystretch{0.8}
	\setlength{\abovecaptionskip}{0cm}
	\centering
	\caption{Ablation study on codebook (\textbf{CB}), siamese decoders (\textbf{SD}) and \textbf{HTA} strategy, performance on both classification and segmentation are reported.}
	\label{ablation_rec}
	\begin{tabular}{lccccc}
		\hline \hline
		\specialrule{0em}{1pt}{1pt}
		\multirow{2}{*}{ID}&\multirow{2}{*}{CB} & \multirow{2}{*}{HTA} & \multirow{2}{*}{SD} & ModelNet40 & ShapeNetPart \\
		&&&& Accuracy & $mIoU_C$\\ \hline
		\specialrule{0em}{1pt}{1pt}
		1&& &           &93.7 & 83.8 \\
		2&& &\checkmark & 93.8 & 84.2\\
		3&\checkmark& &\checkmark & 93.6 & 84.2 \\
		4&\checkmark&\checkmark&  & 93.6 & 84.3 \\
		5&\checkmark&\checkmark&\checkmark &  \textbf{94.1} & \textbf{84.6} \\
		\hline \hline
	\end{tabular}
	\vspace{-10pt}
\end{table}

\begin{table}[]
	\setlength\tabcolsep{3pt}
	\renewcommand\arraystretch{0.8}
	\setlength{\abovecaptionskip}{0cm}
	\centering
	\caption{Ablation study on mask strategy. Performance of different masking types and ratios are investigated, quantified by pre-training loss ($\mathcal{L}_{rec} \times 1000$ and $\mathcal{L}_{align} \times 100$), accuracy on ModelNet40 and $mIoU_C$ on ShapeNetPart.
	}
	\label{ablation_mask}
	\begin{tabular}{lcccccc}
		\hline \hline
		\specialrule{0em}{1pt}{1pt}
		\multirow{2}{*}{Type}&\multirow{2}{*}{Ratio} &  \multirow{2}{*}{$\mathcal{L}_{rec}$}  &  \multirow{2}{*}{$\mathcal{L}_{align}$}  & ModelNet40 & ShapeNetPart \\ 
		&&&& accuracy & $mIoU_C$ \\ \hline
		\specialrule{0em}{1pt}{1pt}
		Random   & 0.45 & 2.38 & 1.00  & 94.0& 84.5 \\
		Random   & 0.55 & 2.40 & 1.21  &94.0& 84.5 \\
		Random   & 0.65 & 2.41 & 1.22  & \textbf{94.1}& \textbf{84.6}  \\
		Random   & 0.75 & 2.44 & 1.50  & 93.7& 84.3 \\
		Random   & 0.85 & 2.47 & 1.73  & 93.1& 84.2\\
		\hline
		\specialrule{0em}{1pt}{1pt}
		Block   & 0.45 &  2.42 &1.80 & 93.2&  84.3\\
		Block & 0.65 &  2.43 & 1.90 & 93.7&84.1 \\
		Block & 0.85 & 2.54 & 1.83 & 93.4&83.9 \\
		\hline \hline
	\end{tabular}
	\vspace{-10pt}
\end{table}

\subsection{Ablation Studies}
\textbf{Point Reconstruction on Codebook. }
To verify the effectiveness of point reconstruction, we first implement the version that reconstructs masked points directly from the prediction of masked representations, which is shown as ID 1 in Table \ref{ablation_rec}. Classification and segmentation results are inferior to those of Point2Vec, indicating that simple addition of point reconstruction is harmful to downstream tasks. Further comparison between ID 2  and ID 3 shows that using a codebook just for point reconstruction demonstrates no obvious performance advantages over that without a codebook. While the HTA strategy can stimulate the function of the codebook, supported by an improvement of 0.4\% $mIoU_C$ on ShapeNetPart (84.6\% vs. 84.2\%) and an improvement of 0.5\% accuracy on ModelNet40 (94.1\% vs. 93.6\%).

\begin{figure}[!hb]
		\setlength{\abovecaptionskip}{0cm}
		\setlength{\belowcaptionskip}{0cm}
		\centering
		\subfigure[$\mathcal{L}_{align}$]{\includegraphics[width=4cm]{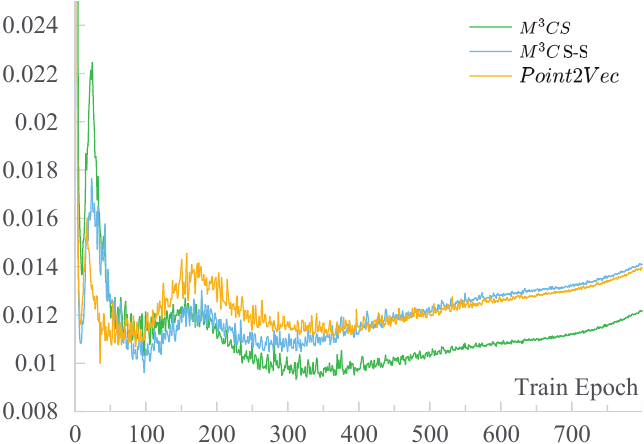}}
		\centering
		\subfigure[$\mathcal{L}_{total}$]{\includegraphics[width=4cm]{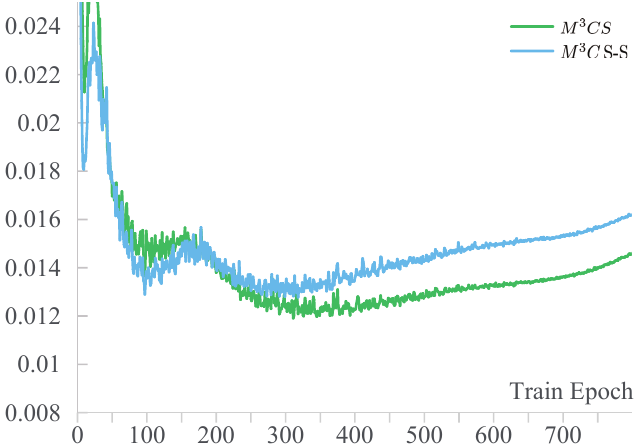}}
	\caption{Comparison of pre-training loss $\mathcal{L}_{align}$ and $\mathcal{L}_{total}$, M$^3$CS-S indicates the version that uses two independent decoders and the masking ratio for all methods is $0.65$.}
	\label{loss}
	\vspace{-5pt}
\end{figure}

\textbf{Siamese Decoders. }
Comparisons between ID 1 and ID 2, ID 4 and ID 5 in Table \ref{ablation_rec} show that siamese decoders can improve downstream fine-tuning under different settings of point reconstruction, as evidenced by improvements of 0.4\% $mIoU_C$ on ShapeNetPart and 0.5\% accuracy on ModelNet40. The performance of fine-tuning demonstrates that multiple reconstruction targets can be coupled to enhance the representation capability of the encoder. 
As shown in Figure~\ref{loss}(a), M$^3$CS obtains a lower $\mathcal{L}_{align}$ in comparison to Point2Vec, which indicates that high-level representation reconstruction can benefit from low-level one by the design of siamese decoders. 
An additional comparison between M$^3$CS and M$^3$CS-S in Figure~\ref{loss}(b) reveals that lower $\mathcal{L}_{total}$ can be achieved with less parameters. 
As discussed above, one decoder can qualify various tasks with different
``prompts'', our siamese decoders create a coupling between high- 
and low-level semantics for more general representations.

\textbf{Masking Strategy. } In this paper, both random and block masking strategies are explored. 
The block masking strategy selects a patch at random and masks it along with its neighbouring patches to achieve a specified masking ratio. 
This strategy results in a more difficult reconstruction task, verified by lower performance in Table \ref{ablation_mask}. 
The optimal masking ratio for random strategy locates at $0.45$-$0.65$ as shown in Table \ref{ablation_mask}, which is slightly smaller than those of Point-MAE and Point2Vec. The typical MAE paradigm employs a higher masking ratio to pass the semantic information from visible patches to masked ones, because it prevents the model from only learning low-level representations of nearby patches \cite{understand_mae}. Point reconstruction in M$^3$CS is performed on the meaningful codebook, 
which is more reliant on semantics. 
As a result, the pre-training of M$^3$CS can adapt to a smaller masking ratio. Notably, the results in Table \ref{ablation_mask} reveal a strong correlation between the pre-training objective and fine-tuning performance; however, reducing the pre-training loss to an extreme value, e.g., $\mathcal{L}_{align}=1.00$, does not result in expected performance. 

\section{Conclusions}
This paper proposed M$^3$CS, a novel multi-target reconstruction method for masked point modeling. Notably, the point reconstruction was performed based on the meaningful codebook. In addition, we introduced siamese decoders with multiple ``prompts'' to exploit the capability of the encoder. M$^3$CS also took advantage of the codebook in downstream fine-tuning. Comprehensive experiments demonstrated that the proposed M$^3$CS can achieve superior performance in various tasks. In the future, a more powerful codebook with an innovative masking strategy will be studied.

{
	\bibliographystyle{m3cs.bst} 
	\bibliography{ref.bib}
}
\end{document}